\documentclass[aps,prx,twocolumn,superscriptaddress]{revtex4-2}
\usepackage[utf8]{inputenc}
\usepackage[T1]{fontenc}% For UTF-8 encoding
\usepackage[english]{babel}
\usepackage{amsmath} % Math symbols and environments
\usepackage{amssymb} % Additional math symbols
\usepackage{amsfonts} % Fonts for math
\usepackage{graphicx} % Include figure files
\usepackage{mathrsfs} % Math script fonts
\usepackage{bm} % Bold math symbols
\usepackage{color} % For colored text (e.g., comments)
\usepackage{hyperref} % Hyperlinks
\hypersetup{colorlinks=true,linkcolor=blue,citecolor=cyan} % Hyperlink customization
\usepackage{physics} % Physics symbols like bra-ket
\usepackage{braket} % For bra-ket notation
\usepackage{bbold} % For bold 1 (identity matrix)
\usepackage{siunitx} % For SI units
\usepackage{xcolor} % For advanced color options
\usepackage{times} % Use Times font
\usepackage{caption}
\usepackage{ragged2e}
\usepackage{physics}
\usepackage{orcidlink}

\usepackage{float}
\usepackage{physics}
\DeclareUnicodeCharacter{0301}{'}

\def\ii{\text{i}}
\def\ee{\operatorname{e}}

\newcommand{\pair}[2]{\left\langle #1,#2 \right\rangle}
\newcommand{\change}[1]{\textcolor{black}{#1}}

\begin{document}

\title{Near-Equilibrium Propagation training in nonlinear wave systems}

\author{Karol Sajnok\orcidlink{0009-0004-5899-8923}}
\email{ksajnok@cft.edu.pl}
\affiliation{Center for Theoretical Physics, Polish Academy of Sciences, Aleja Lotników 32/46, 02-668 Warsaw, Poland}
%\affiliation{Institute of Theoretical Physics, University of Warsaw, ul. Pasteura 5, 02-093 Warsaw, Poland}

\author{Michał Matuszewski\orcidlink{0000-0001-8830-3302}}
\email{mmatuszewski@cft.edu.pl}
\affiliation{Center for Theoretical Physics, Polish Academy of Sciences, Aleja Lotników 32/46, 02-668 Warsaw, Poland}
\affiliation{Institute of Physics, Polish Academy of Sciences, Aleja Lotników 32/46, 02-668 Warsaw, Poland}

%%%%%%%%%%%%%%%%%%%%%%%%%%%%%%%%%%%%%%%%%%%%%%%%
\date{\today}
\begin{abstract}
Backpropagation learning algorithm, the workhorse of modern artificial intelligence, is notoriously difficult to implement in physical neural networks. Equilibrium Propagation (EP) is an alternative with comparable efficiency and strong potential for in-situ training. We extend EP learning to both discrete and continuous complex-valued wave systems. In contrast to previous EP implementations, our scheme is valid in the weakly dissipative regime, and readily applicable to a wide range of physical settings, even without well defined nodes, where trainable inter-node connections can be replaced by trainable local potential. We test the method in driven–dissipative exciton-polariton condensates governed by generalized Gross–Pitaevskii dynamics. Numerical studies on standard benchmarks, including a simple logical task and handwritten-digit recognition, demonstrate stable convergence, establishing a practical route to in-situ learning in physical systems in which system control is restricted to local parameters.
\end{abstract}

\maketitle

\section{Introduction}
Neuromorphic learning in physical substrates promises orders-of-magnitude gains in throughput and energy efficiency by exploiting native dynamics for training and inference~\cite{Markovic2020Physics,sangwan2020neuromorphic,Grollier2020Spintronics,kudithipudi2025neuromorphic,shastri2021photonics,matuszewski2024role}. Optical and solid-state platforms provide reconfigurable couplings with fast readout, yet end-to-end \emph{in situ} training remains challenging because backpropagation assumes an explicit computational graph and accurate adjoint dynamics, assumptions fragile in analog hardware with hidden couplings, latency, and dissipation~\cite{Markovic2020Physics,momeni2025training}. Driven–dissipative wave systems, including photonic neural networks, are especially attractive due to their throughput and scalability~\cite{mcmahon2023physics,wetzstein2020inference,shastri2021photonics,matuszewski2024role,momeni2025training}. While {\it in silico} training can work for small devices, model–experiment mismatch in larger systems motivates efficient \emph{in situ} alternatives~\cite{hughes2018training,nakajima2022physical,wright2022deep,oguz2023forward,pai2023experimentally,momeni2023backpropagation}. However, nonlinear complex-valued dynamics couple amplitude and phase to gain and loss, complicating accurate gradient estimation.

Equilibrium Propagation (EP) is an energy-based two-phase algorithm that estimates gradients locally from contrasts between free and nudged steady states~\cite{ScellierBengio2017}. The system first relaxes under the input, then under weak output-dependent nudging; their steady-state difference yields a local update rule. EP is equivalent to recurrent backpropagation~\cite{Scellier2019Equivalence,Ernoult2019Updates} and achieves high accuracy on large-scale digital \change{benchmarks~\cite{Ernoult2019Updates,laborieux2021scaling,kerjan2026training}.} It has been extended to convolutional, spiking, oscillator, \change{thermodynamic}, and quantum networks, as well as to Lagrangian and agnostic learning frameworks~\cite{laborieux2021scaling,Martin2021EqSpike,ricco2024training,rageau2025training,wang2024oscillators,zoppo2022iscas,scellier2024quantum,massar2025equilibrium,WanjuraMarquardt2025QEP,scellier2021PhD,Massar2025LagrangianEP,scellier2022agnostic,lipka2026thermodynamic}. \change{EP was demonstrated experimentally in mechanical~\cite{altman2024experimental}, resistor~\cite{dillavou2022demonstration,dillavou2024machine} and memristor~\cite{yi2023activity} networks. It was also partially implemented experimentally on a D-Wave quantum circuit~\cite{Laydevant2024IsingEP} and optical coherent Ising machines~\cite{abeele2026optical,fan2026optimizing}. }

\begin{figure}
    \centering
    \includegraphics[width=0.48\textwidth]{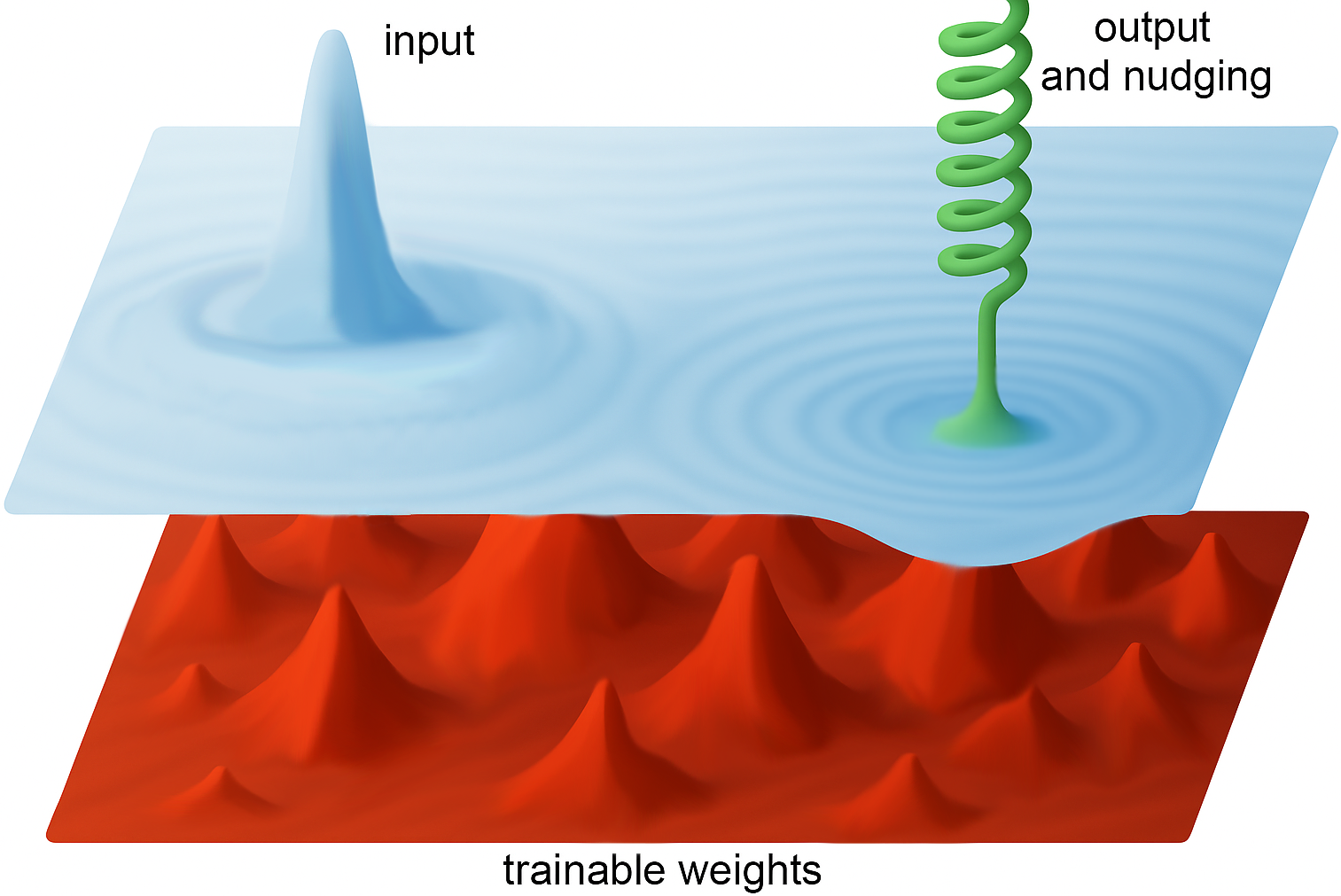}
        \caption{\justifying Scheme of NEP implementation in a continous wave system (light blue) with input drive, output  and nudging force (green spring) and trainable potential parameters (red).}
    \label{fig:eqprop}
\end{figure}

Here, we introduce Near-Equilibrium Propagation (NEP), a generalization of EP to driven–dissipative complex-wave systems (Fig.~\ref{fig:eqprop}). 
Our approach extends previous results in several ways: (i) operates in the near-equilibrium regime of weak pumping and dissipation; (ii) is applicable to oscillatory and wave systems, extending beyond conventional EP based on energy relaxation; (iii) can train \emph{local parameters} instead of inter-node weights, thus covering a broader class of physical systems; and (iv) accommodates both discrete and continuous complex-valued fields. Some of these aspects were partially explored before. Generalization of EP to real-valued vector-field dynamics without a well-defined energy function was considered in~\cite{Scellier2018VectorField}. Training oscillatory systems with EP was studied in~\cite{rageau2025training,wang2024oscillators,zoppo2022iscas}, but those models assumed energy-gradient descent rather than Hamiltonian dynamics, yielding relaxation instead of oscillatory behavior. EP in complex networks was introduced in~\cite{laborieux2022holomorphic} for holomorphic functionals, which do not correspond to physical real-valued quantities like energy or Hamiltonian. In~\cite{hughes2019wave}, the analogy between continuous wave systems and neural networks was recognized, but no method for in-situ training was provided, relying instead on backpropagation in a discretized digital model.

To test the framework, we model NEP learning in exciton-polariton condensates, a platform combining light and matter~\cite{byrnes2014exciton,sanvitto2016road,fraser2016physics}. Polaritonic neural networks have been explored in several architectures~\cite{opala2019neuromorphic,ballarini2020polaritonic,Mirek2021Binarized,kavokin2022polariton,opala2023harnessing,tyszka2023lif,opala2022training,opala2024room}, and recent work highlights their energy efficiency for inference~\cite{matuszewski2021energy,matuszewski2024role}. We demonstrate stable NEP training on XOR and MNIST tasks, derive local update rules from free–nudged contrasts, and model a realistic implementation using the driven–dissipative Gross–Pitaevskii equation, including pumping geometry and output nudging. Numerical results confirm convergence and indicate experimental feasibility.

\section{Near-Equilibrium Propagation in complex-valued systems.}
We define the NEP learning protocol for driven–dissipative systems. An oscillatory or wave system is described by a time-dependent complex field $\psi(\vec{r},t)$ defined over a spatial domain $\vec{r}\in\mathcal{M}$, which may be continuous or discrete depending on the physical realization.

Training proceeds via a two-step procedure for each input sample $\text{X}=(x_1,x_2,\dots)$.
In the first, free-evolution phase, the system follows
\begin{align} \label{eq:free}
\partial_t \psi(\vec{r},t) &= \kappa_\psi (\psi(\vec{r},t),\theta,\text{X}),
\end{align}
where $\theta=(\theta_1,\theta_2,\dots)$ is the vector of trainable parameters, such as local potentials, inter-node couplings, or input weights. Unlike in conventional EP, Eq.~\eqref{eq:free} need not derive from a variational principle with an energy functional. We assume that after sufficiently long evolution, the system reaches an oscillating steady state $\psi(\vec{r},t)=\Psi_0(\vec{r})\ee^{i\omega t}$, where $\omega$ denotes frequency, implying $\kappa_0(\Psi_0(\vec{r}),\theta,\text{X}):=\kappa_\psi\ee^{-i\omega t}=i\omega\Psi_0(\vec{r})$. We assume $\omega$ is locked to a resonant drive, $\omega=\omega_{\rm D}={\rm const}$, independent of $\theta$ and $\text{X}$. Moving to the rotating frame $\Psi(\vec{r},t)=\psi(\vec{r},t)\ee^{-i\omega_{\rm D} t}$ gives
\begin{equation}
\label{eq:free_rot}
\partial_t\Psi(\vec{r},t)=\kappa(\Psi(\vec{r},t),\theta,\text{X})\overset{\mathrm{s.s.}}{=}0,
\end{equation}
with $\kappa=\kappa_0(\Psi,\theta,\text{X})-i\omega_{\rm D}\Psi$, where the last equality in~\eqref{eq:free_rot} is fulfilled in the steady state $\Psi(\vec{r},t)=\Psi_0(\vec{r})$.

In the second, nudged phase, the system is perturbed with respect to the target. The goal is to tune $\theta$ to minimize the cost function $C(\Psi_0)$, quantifying output–target discrepancy. To achieve this, we add a driving term proportional to the cost gradient
\begin{align} \label{eq:kappa_beta}
\partial_t \Psi &= \kappa - \ii\beta\pdv{ C(\Psi_0,\overline\Psi_0)}{\overline \Psi_0} =: \kappa^\beta,
\end{align}
where $\beta$ is a small real parameter and $\pdv{C}{\overline\Psi}$ is a Wirtinger derivative. The perturbation includes an imaginary factor, and derivatives are taken with respect to the conjugate variable. This yields a simple in-situ gradient-descent rule. We assume the system relaxes to a new, nudged steady state satisfying $\kappa^\beta(\Psi,\theta,\text{X},\beta)\overset{\mathrm{s.s.}}{=}0$.

For complex-field systems, the output can depend on both amplitude and phase; here, we focus on amplitude. For instance, the mean-squared error (MSE) cost in the continuous case is
\begin{align} \label{eq:mse}
C_{\rm MSE}(\Psi_0,\overline\Psi_0) &= \frac{1}{2} \int \limits_\mathcal{Y} \left(|\Psi_\text{Y}|^2 - |\Psi_0(\vec{r})|^2\right)^2 d\vec{r},
\end{align}
where $\mathcal{Y}$ is the output region. This choice leads, via Eq.~\eqref{eq:kappa_beta}, to the evolution equation
\begin{align} \label{eq:driven_psi_mse}
\kappa^\beta_{\rm MSE} &= \kappa + \ii\beta \Theta_{\mathcal{Y}} \left(|\Psi_\text{Y}|^2 - |\Psi_0|^2\right)\Psi_0,
\end{align}
with Heaviside step function $\Theta_{\mathcal{Y}}(\vec{r})=1$ for $\vec{r}\in\mathcal{Y}$ and $0$ otherwise. Another example is the categorical cross-entropy (CCE) cost, useful for multi-class tasks, given in the discrete case by
\begin{align} \label{eq:cce}
C_{\rm CCE}(\Psi_0,\overline\Psi_0) &= - \sum_i |\Psi_\text{Y}(\vec{r_i})|^2 \ln \sigma(|\Psi_0(\vec{r_i})|^2),
\end{align}
where $\sigma(|\Psi(\vec{r_i})|^2)=\frac{\exp(|\Psi(\vec{r_i})|^2)}{\sum_j \exp(|\Psi(\vec{r_j})|^2)}$ is the softmax function, and the sum runs over all output nodes $\vec{r}_i\in\mathcal{Y}$ corresponding to output classes. The nudged evolution then becomes
\begin{align} \label{eq:driven_psi_cce}
\kappa^\beta_{\rm CCE} &= \kappa + \ii\beta \sum_i \left(|\Psi_\text{Y}(\vec{r_i})|^2 - \sigma(|\Psi_0(\vec{r_i})|^2) \right)\Psi_0(\vec{r_i}).
\end{align}
Thus, the nudged perturbation depends on the chosen cost function and is confined to $\mathcal{Y}$.

Parameter updates $\theta$ can be applied per sample, batch, or epoch. They minimize the cost along the steepest gradient
\begin{align} \label{eq:delta_theta}
\Delta \theta &\propto - \left.\pdv{C}{\theta}\right|_{\beta=0} = \ii \pair{ \pdv{\overline\Psi}{\beta} }{ \pdv{\overline\kappa}{\theta} } - \ii \pair{ \pdv{\Psi}{\beta} }{ \pdv{\kappa}{\theta} },
\end{align}
where $\pair{f}{g}$ is the inner product, $\pair{f}{g}:=\int_{\mathcal M}\dd{\vec{r}}\overline{f}(\vec{r})g(\vec{r})$ in the continuous case, and $\pair{f}{g}:=\sum_{i\in\mathcal M}\overline{f}(\vec{r}_i)g(\vec{r}_i)$ in the discrete one. The derivations for both cases are given in Appendix~\ref{appA}\change{, while Appendix~\ref{appB} provides a detailed comparison with the results of~\cite{Scellier2018VectorField}}. 

NEP training generalizes~\cite{Scellier2018VectorField} and holds for any complex wave equation, provided derivatives of $\kappa(\Psi,\overline\Psi,\theta,\text{X})$ satisfy near-equilibrium conditions
\begin{align}
\pdv{\kappa(\vec{r})}{\overline\Psi(\vec{s})} = \pdv{\kappa(\vec{s})}{\overline\Psi(\vec{r})}, \qquad
\pdv{\kappa(\vec{r})}{ \Psi(\vec{s})} \approx - \pdv{ \overline \kappa(\vec{s})}{ \overline\Psi(\vec{r})}. \label{eq:condition}
\end{align}
Here, {\it equilibrium} is understood more broadly than in previous EP works~\cite{ScellierBengio2017,wang2024oscillators,rageau2025training}, referring to a stationary fixed point of Eq.~\eqref{eq:free_rot}. When a Hamiltonian functional $H(\Psi,\overline\Psi)$ exists such that \change{$ \kappa(\vec{r})\approx \{\Psi(\vec{r}),H\}=-\ii\pdv{H}{\overline\Psi(\vec{r})}$ with $\{\cdot\}$ as the Poisson bracket, Eq.~\eqref{eq:condition} is naturally satisfied as a consequence of Clairaut's theorem}. As shown below, even when not exactly fulfilled, training can still converge successfully.

During inference, nudging $\beta$ is switched off, and outputs are read directly from the steady-state amplitudes $\Psi_\text{Y}$ in the output region, with the system driven by input $X$ as in the free phase.

\section{Implementation in exciton-polariton systems.}
\begin{figure}
\centering
\includegraphics[width=0.48\textwidth]{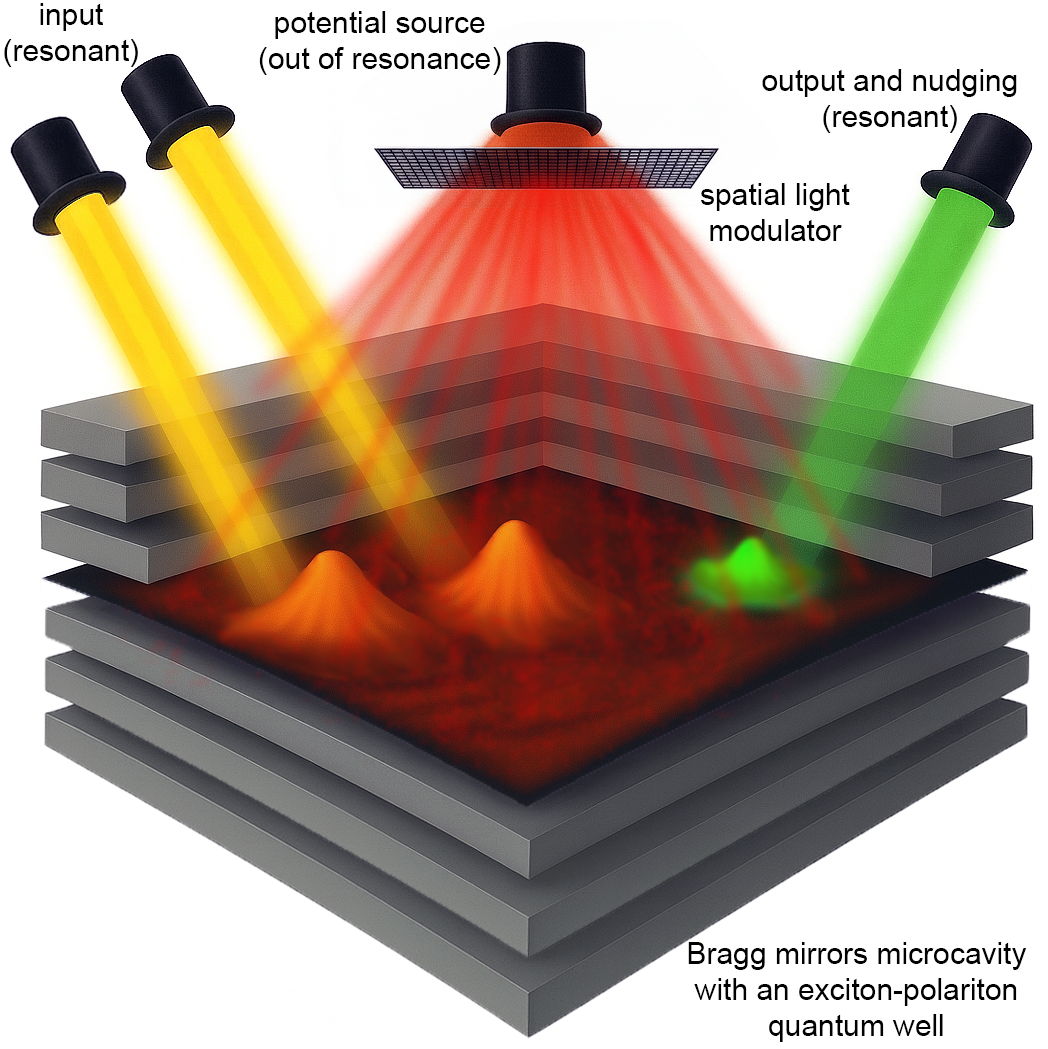}
\caption{\justifying Example of a polaritonic system with two inputs and one output. Two localized laser beams (yellow) encode the inputs. A spatially modulated trainable potential (red) is applied via a SLM. The selected area (green) marks the output region where optical nudging and readout occur. Input and output beams share the same frequency $\omega_D$.}
\label{fig:eqprop_polaritons}
\end{figure}

Exciton-polariton systems are ideal for optical neural networks owing to their strong nonlinearity, ultrafast dynamics, and high energy efficiency~\cite{matuszewski2021energy,matuszewski2024role,ballarini2020polaritonic,Mirek2021Binarized}. The NEP setup, shown in Fig.~\ref{fig:eqprop_polaritons}, consists of a cavity sample hosting polaritons, resonant laser beams for input pumping and output nudging, and a non-resonant laser forming the potential. Trainable parameters include both software input weights and the physical potential landscape controlled by a spatial light modulator (SLM).

For illustration, we adapt the general NEP framework to driven–dissipative polaritons described by a generalized continuous GPE with dissipation $\gamma$, complex potential $V(\vec{r})$, resonant pumping $P(\vec{r})$, and nonlinear function $f(\Psi)$—either density response $f_{\rm d}(\Psi)=g_{\rm d}|\Psi|^2$ or saturation response $f_{\rm s}(\Psi)=\frac{g_{\rm s}}{1+|\Psi|^2}$. In the rotating frame, the polariton wavefunction $\Psi(\vec{r},t)$ evolves as
\begin{align} \label{eq:free_gpe}
\partial_t \Psi \equiv \kappa &= -\frac{\ii}{\hbar}\left(-\frac{ \hbar^2}{2m}\nabla^2 + V+f(\Psi) -\ii\gamma \right)\Psi + P,
\end{align}
where $P(\vec{r})=\sum_k w_k X_k G(\vec{r}-\vec{r}_k)$, with complex input weights $w_k$ for inputs $X_k$ and normalized Gaussian envelopes \change{$G(\vec{r}-\vec{r}_k)=\frac{1}{W\sqrt{2 \pi }}\exp\left[-\frac{(\vec{r}-\vec{r}_k)^2}{2W^2}\right]$} of width $W$ centered at $\vec{r}_k$.
The complex potential $V(\vec{r})$ and pumping weights $w_k$ constitute the trainable parameters $\theta$. Although the terms $P$, $\gamma$, and $\operatorname{Im}V$ break Hamiltonian dynamics, we show below that NEP remains accurate when $\gamma$ and $\operatorname{Im}V$ are small compared to the other terms (the near-equilibrium regime). The Laplacian term, being parity-symmetric, satisfies both conditions in Eqs.~\eqref{eq:condition}.

In the nudged evolution phase, the perturbation is introduced in the output region $\mathcal{Y}$, as defined in Eq.~\eqref{eq:mse}, depending on the chosen cost function. Experimentally, this corresponds to applying a resonant pump in $\mathcal{Y}$ with amplitude proportional to the error and a $\frac{\pi}{2}$ phase shift relative to the phase of the steady state in the output region. The steady state $\Psi_0(\vec{r})$ satisfies $\kappa^\beta(\Psi^\beta,V,w,P,\beta)=0$.

To determine update rules for the two training parameters—the potential $V(\vec{r})$ and pumping weights $w(\vec r_k)$—we employ the NEP rule Eq.~\eqref{eq:delta_theta}. Using the derivative of $\kappa$ with respect to the real potential $V(\vec{r})$ gives
\begin{align}
\Delta V(\vec{r}) \propto - \pdv{ C}{ V(\vec{r})} &= -\frac{1}{\hbar} \pdv{|\Psi_0(\vec{r})|^2}{\beta}. \label{eq:delta_V_polariton}
\end{align}
Analogously, for the real pump weights,
\begin{align}
\Delta w_k \propto - \pdv{C}{w_k} &= - 2 \int \dd \vec{r} X_k G(\vec{r}-\vec{r}_k) \operatorname{Im} \pdv{\Psi_0(\vec{r})}{\beta}. \label{eq:delta_w_polariton}
\end{align}
Thus, updates for both the potential and pump weights depend on local steady-state fields $\Psi_0(\vec{r},\beta)$ obtained in the free and driven phases. The parameter $\beta$ controls the nudging strength and must be small enough for accurate derivative estimation at $\beta=0$, yet large enough to induce a measurable change in the wavefunction.

\section{Numerical results.}
In simulations, we use a discretized version of the GPE and implement the two-phase NEP algorithm on a finite spatial grid. In dimensionless units, the discretized GPE reads
\begin{align}\label{eq:discrete_gpe}
\kappa_i &= \frac{\ii}{2}(\Psi_{i+1}-2\Psi_i+\Psi_{i-1})-\ii(V_i+f(\Psi_i)-\ii\gamma)\Psi_i + P_i.
\end{align}
Dirichlet boundary conditions are imposed, realizable experimentally via a steep out-of-resonant potential wall, though the framework is independent of boundary choice. The evolution equation is integrated using a fourth-order Runge–Kutta (RK4) scheme.

\begin{figure}
\centering
\includegraphics[width=0.48\textwidth]{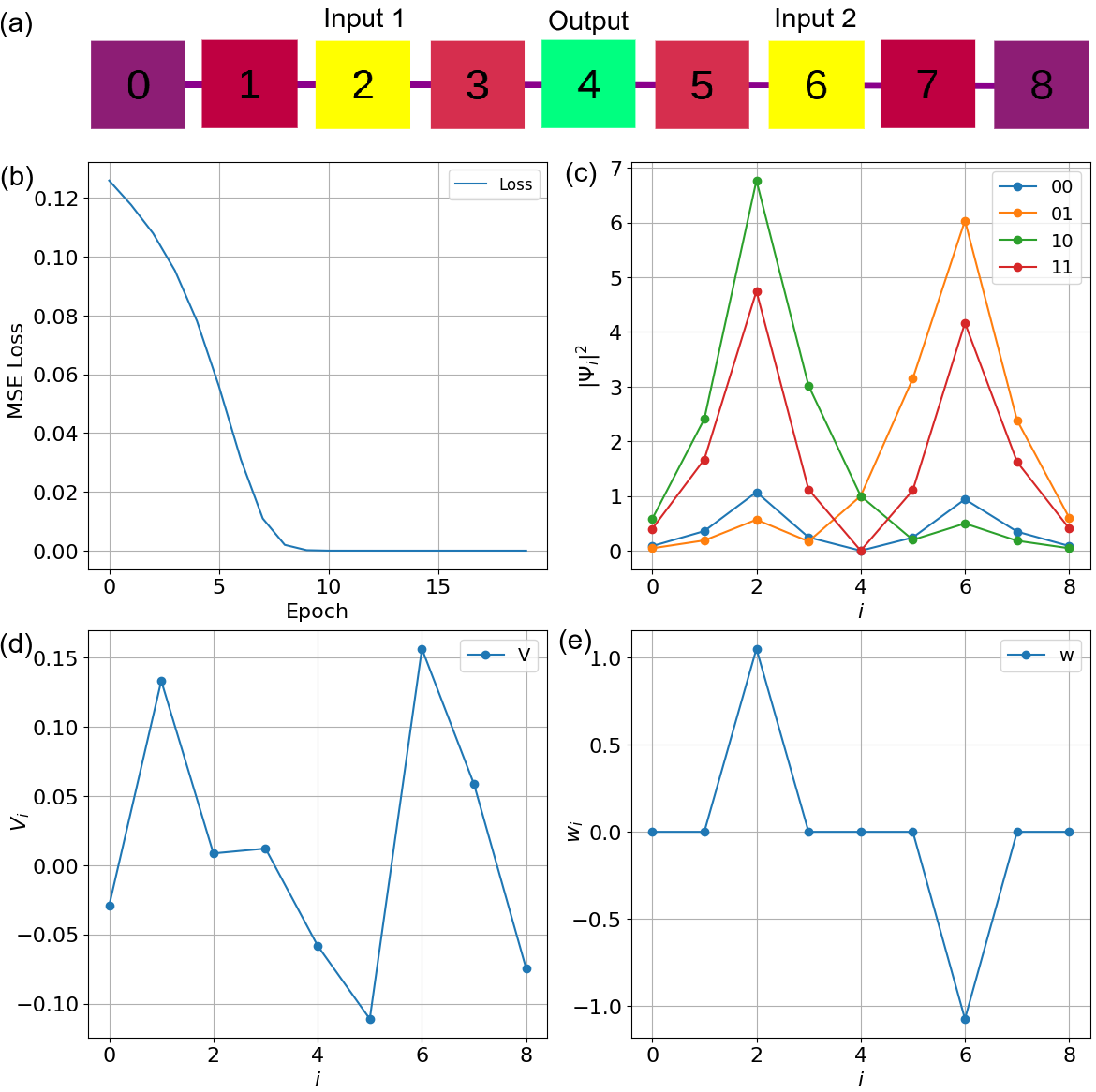}
\caption{\justifying Training a 9-node 1D network for the 2-input XOR task. Inputs are applied symmetrically at sites 2 and 6, and the output is read at site 4 (a). Both potential $V$ and pump weights $w$ are trained. (b) Training loss; (c) steady-state field magnitudes $|\Psi_i|^2$ for each input; (e) trained potential $V$; (d) pump weights $w$. Final outputs: $|\Psi_4|^2=(0.00,0.92,1.06,0.01)$ for inputs $(00,01,10,11)$.}
\label{fig:xor}
\end{figure}

As a proof of principle, we train a one-dimensional polariton network to implement the two-input XOR function. The simulation employs a 9-node grid, with two input nodes and one output node $\mathcal{Y}$. Figure~\ref{fig:xor} shows the system geometry and optimization results. Input samples $\text{X}$ correspond to four binary combinations $(00, 01, 10, 11)$ and are encoded as pump amplitudes $P_i= w_i \left( \delta_i^{i_1} X_1 + \delta_i^{i_2} X_2 \right)$, where $w_i$ are real input weights, and $X_k$ ($k=1,2$) are the inputs located at sites $i_k$. The target pattern $\Psi_\text{Y}$ requires high amplitude ($|\Psi_\text{Y}|^2=1$) for $(01,10)$ and low ($|\Psi_\text{Y}|^2=0$) otherwise. The cost function is the MSE between output intensity and target intensity for each input.

As shown in Fig.~\ref{fig:xor}(b), convergence occurs after about 10 epochs with nudging parameter $\beta=0.01$. Both potential $V$ and pump weights $w$ are trained following Eqs.~\eqref{eq:delta_V_polariton}–\eqref{eq:delta_w_polariton}, using learning rates ${\rm lr}_V={\rm lr}_w=0.1$. The GPE parameters are $g=0.1$, $\gamma=0.1$, and nonlinear saturation response $f_{\rm s}(\Psi_i)=\frac{g_{\rm s}}{1+|\Psi_i|^2}$. Other panels of Fig.~\ref{fig:xor} show steady-state amplitudes, potential, and input weights in the trained system. Appendix~B demonstrates that optimization remains effective regardless of input/output node positions and can rely on training both $V_i$ and $w_i$, or only $V_i$ with fixed $w_i$. The method is robust to random potential perturbations—e.g., structural inhomogeneities in experimental samples—verified by training with fixed small random fluctuations, which did not affect convergence.
 
\begin{figure}
\centering
\includegraphics[width=0.48\textwidth]{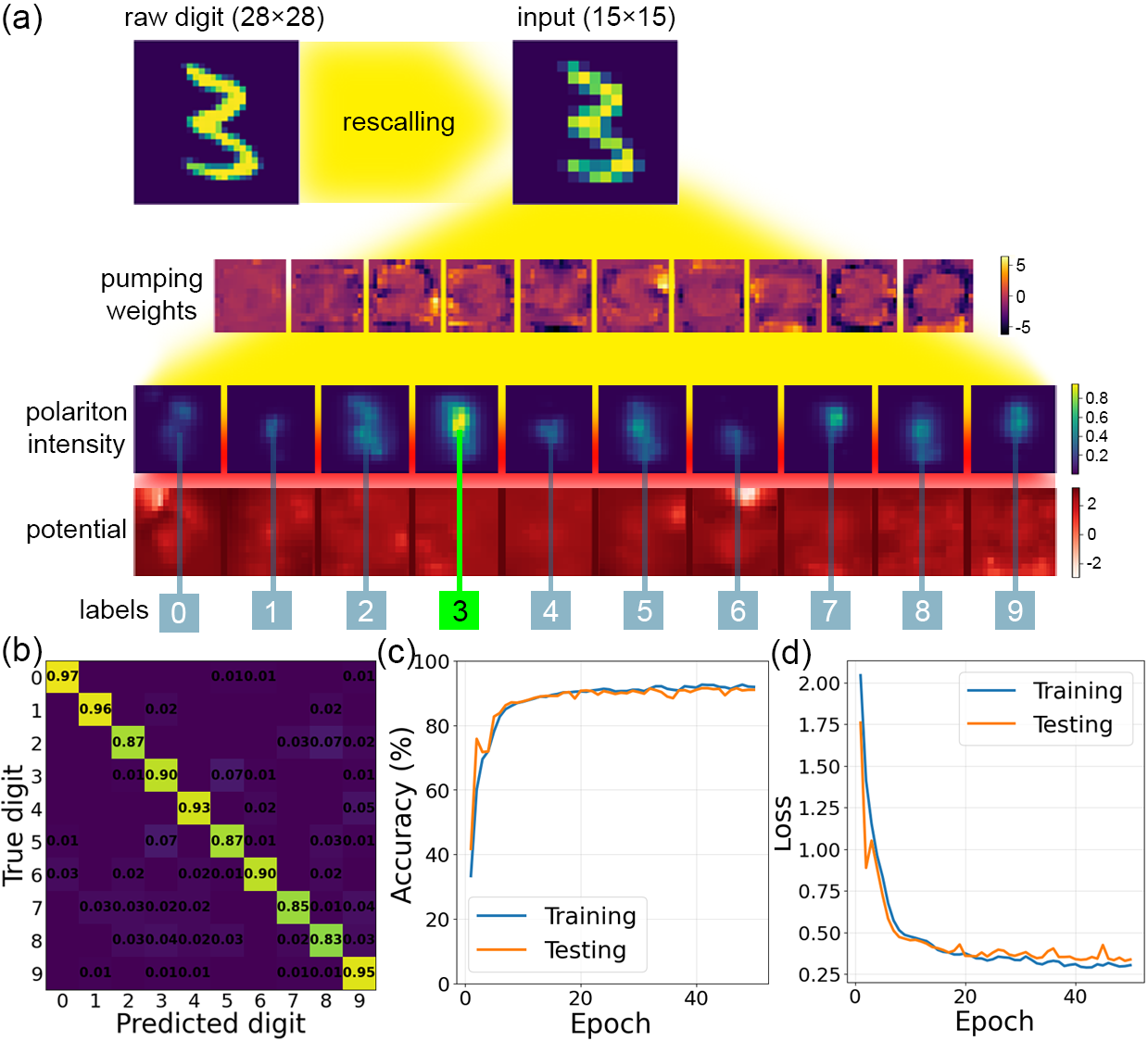}
\caption{\justifying Training a $15 \times 150$ 2D network to classify handwritten MNIST digits. Each of the ten $15 \times 15$ cells receives a rescaled input image. The output region $\mathcal{Y}$ contains ten central nodes corresponding to digits $0$–$9$. Panel (a): data flow, including rescaling, replication, optical injection with pumping weights, nonresonant potential, final polariton intensity. The cells are separated by a blocking potential (dark red). Panel (b): confusion matrix with final test-set accuracies. Panels (c,d): training and testing accuracy and loss.
}
\label{fig:mnist}
\end{figure}

To demonstrate the capability of the NEP algorithm to train multi-class classification tasks, we trained a 2D discrete polariton network to classify the MNIST handwritten digits dataset ($1000$ samples per digit), see Fig.~\ref{fig:mnist}. The simulation is performed on a $15 \times 150$ grid, where the input region $\mathcal{X}$ corresponds to the entire network area, with each of ten $15 \times 15$ cells receiving a copy of the same input. Input patterns are generated by resizing and normalizing MNIST images. The output region $\mathcal{Y}$ consists of ten central nodes, each representing one class. The output pattern $\Psi_\text{Y}$ is one-hot encoded, with each output node required to exhibit high intensity ($\sigma(|\Psi_\text{Y}|^2)=1$) for the correct digit and low intensity ($\sigma(|\Psi_\text{Y}|^2)=0$) for all others, with the cost function given by Eq.~\eqref{eq:cce}. Convergence to a test accuracy of $\approx 90.3\%$ is achieved after about 20 training epochs with nudging parameter $\beta=0.01$, learning rates ${\rm lr}_V = {\rm lr}_w = 0.1$, resonant pumping amplitude normalized to unity, GPE parameters $g=0.04$, $\gamma=0.07$, and density-response nonlinearity $f_{\rm d}(\Psi_i)=|\Psi_i|^2$. Notably, the pumping weights shown in Fig.~\ref{fig:mnist} form patterns that resemble handwritten digits, recognizable to the human eye.

\begin{figure}
\centering
\includegraphics[width=0.48\textwidth]{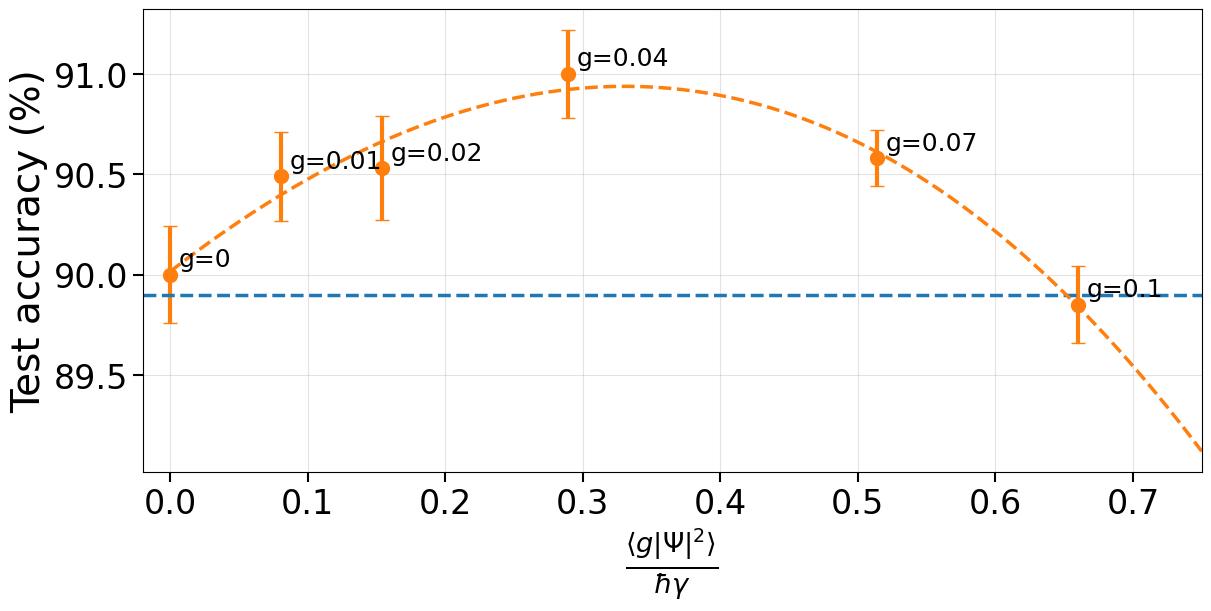}
\caption{\justifying Mean test set accuracy over the last $10$ out of 50 epochs for the setup of Fig.~\ref{fig:mnist}, plotted as a function of the mean nonlinearity-to-loss ratio. The dashed orange line shows a third-order polynomial fit, whereas the dashed blue line marks the accuracy of a fully connected linear network trained by backpropagation on the same inputs.}
\label{fig:mnist-g}
\end{figure}

Varying the nonlinearity $g$ reveals optimal performance for a moderate value $g=0.04$, corresponding to the mean nonlinearty-to-loss ratio $\frac{\langle g|\Psi|^2\rangle}{\hbar \gamma}\approx0.3$, which yields the lowest final loss and highest validation accuracy \change{of $\approx 91\%$}, see Fig.~\ref{fig:mnist-g}. Larger values of $g$ tend to destabilize learning. Training only the potential, without optimizing the pumping weights, reduces performance but confirms that optimization can proceed via purely optical, rather than electronic, updates (Appendix~\ref{appC}). A benchmark fully connected linear network with no hidden neurons and 10 outputs, trained by backpropagation, achieves $\approx 89.9\%$ accuracy on the same MNIST test subset. Previous studies reported that oscillatory networks with local connections perform poorly on MNIST~\cite{wang2024oscillators}. We attribute the improved performance to nonequilibrium wave dynamics, which overcome the locality of node couplings.

\change{Appendix~\ref{appC} presents additional simulation results for the XOR and MNIST tasks, demonstrating the robustness of NEP to restricted parameter training and variations in network architecture.}

\section{Experimental outlook and summary.}\label{Sec:Summary}
The NEP protocol can be experimentally realized with current polariton systems. Numerical simulations for the MNIST task ($1000$ samples per digit) converge after about $20$ epochs for $\gamma = 0.1$, requiring two steady-state measurements per sample, each with $\approx5000$ RK4 steps and $\dd{t}=0.1$. In physical units, the full training procedure completes within $T_{\rm phys} \approx \SIrange{0.1}{10}{\milli\second}$ for typical $\gamma^{-1} \approx \SIrange{1}{100}{\pico\second}$~\cite{deng2010exciton,byrnes2014exciton}, nearly six orders of magnitude faster than GPU-based numerical integration (NVIDIA GeForce RTX 2080 Ti, 11 GB). Thus, NEP provides both conceptual and substantial practical advantages. For GaAs-based microcavity~\cite{estrecho2019direct}, a typical linewidth $\hbar\gamma\approx\SI{0.1}{\milli\electronvolt}$ and interaction constant $g=\SI{1}{\micro\electronvolt\,\micro\meter\squared}$ \cite{estrecho2019direct}, the numerical parameters used in the simulations shown in Fig.~\ref{fig:mnist} correspond to exciton density $|\Psi|^2\approx\SI{30}{\micro\meter^{-2}}$. The trained parameters correspond to coupling energy of $\approx\SI{1}{\milli\electronvolt}$, optically induced potential of $\approx\SI{0.4}{\milli\electronvolt}$ and resonant pumping amplitude $\approx\SI{0.8}{\milli\electronvolt}$, all well within experimentally accessible range.

We have presented the NEP scheme, which transforms a nonlinear wave system into a machine-learning device trained via steady-state contrasts. Unlike previous approaches, our algorithm operates near equilibrium, trains local parameters instead of inter-node weights, and applies to both discrete and continuous systems. The protocol uses standard optical tools—SLMs and camera-based readouts—is robust to moderate inhomogeneities, and naturally supports tiling into weakly coupled cells for large-scale tasks. Altogether, these features make NEP a realistic framework for laboratory implementation, enabling in-situ, ultrafast, energy-efficient training on polaritonic hardware within experimental reach.

After the submission of our work, Scurria \emph{et al.} reported an EP scheme for nonequilibrium steady states formulated in a real-valued state space~\cite{scurria2026equilibrium}, whereas our NEP targets driven--dissipative polariton networks with complex fields and implements nudging via Wirtinger derivatives.

\vspace{1em}

\section*{Data availability.}
The data that support the findings of this article are openly available \cite{zenodo}.

\section*{Acknowledgments.} We thank Andrzej Opala for useful discussions. The authors acknowledge support from the European Union EIC Pathfinder project ‘Neuromorphic Polariton Accelerator’ (PolArt, ID 101130304) and from the National Science Center, Poland grant 2021/43/B/ST3/00752.

\bibliography{bibliography}

\appendix

\section{Derivation of the Update Rule} \label{appA}
\textit{Continuous variable case.} We derive the update rule for a trainable parameter $\theta(\vec{r})$ in an arbitrary complex wave system. Using the chain rule for functional derivatives, the derivative of the cost function $C(\Psi,\overline\Psi)$ with respect to $\theta(\vec{r})$ reads
\begin{align} \label{eq:app_delta_V1}
\pdv{ C}{ \theta(\vec{r})} &= \int \dd\vec{s} \left( \pdv{C}{\Psi(\vec{s})} \pdv{\Psi(\vec{s})}{\theta(\vec{r})} + \pdv{ C}{ \overline\Psi(\vec{s})} \pdv{\overline\Psi(\vec{s})}{\theta(\vec{r})} \right),
\end{align}
where integration is over the spatial domain $\mathcal{M}$, and $\Psi=\Psi(\theta,\overline\theta)$, $\overline\Psi=\overline\Psi(\theta,\overline\theta)$ are steady-state fields depending implicitly on $\theta$ and $\overline\theta$.

From Eq.~\eqref{eq:kappa_beta},
\begin{align}
\pdv{ C}{ \Psi} &= -\ii\pdv{ \overline \kappa^\beta}{ \beta}, \label{eq:app_cost_diff2}
\end{align}
so Eq.~\eqref{eq:app_delta_V1} becomes
\begin{align} \label{eq:app_delta_V2}
\pdv{ C}{ \theta(\vec{r})} &=- \ii\int \dd\vec{s} \left(\pdv{ \overline\kappa^\beta}{ \beta} \pdv{\Psi(\vec{s})}{\theta(\vec{r})} - \pdv{\kappa^\beta}{ \beta} \pdv{\overline\Psi(\vec{s})}{\theta(\vec{r})} \right),
\end{align}
since $\pdv{C}{\overline\Psi}=\overline{\pdv{C}{\Psi}}$ for real-valued $C$.

In the nudged steady state, $\kappa^\beta(\Psi^\beta,\overline{\Psi}^\beta,\theta,\overline{\theta},\beta)=0$, where $\Psi^\beta$ and $\overline{\Psi}^\beta$ depend explicitly on $\theta,\overline{\theta},\beta$. Treating these as independent variables and differentiating with respect to $\beta$ yields
\begin{align}
\pdv{ \kappa^\beta(\vec{s})}{\beta} &= - \int \dd\vec{u} \left( \pdv{\kappa^\beta(\vec{s})}{\Psi^\beta(\vec{u})} \pdv{\Psi^\beta(\vec{u})}{\beta} +\pdv{\kappa^\beta(\vec{s})}{\overline\Psi^\beta(\vec{u})} \pdv{\overline\Psi^\beta(\vec{u})}{\beta} \right),\label{eq:app_stat_cond}
\end{align}
evaluated at $\beta=0$.

Similarly, differentiating the stationary condition with respect to $\theta(\vec{r})$ gives
\begin{align} \label{eq:app_stat_cond2}
\pdv{ \kappa^\beta(\vec{u})}{\theta(\vec{r})} &= - \int \dd\vec{s} \left( \pdv{\kappa^\beta(\vec{u})}{\Psi^\beta(\vec{s})} \pdv{\Psi(\vec{s})}{\theta(\vec{r})} +\pdv{\kappa^\beta(\vec{u})}{\overline{\Psi}^\beta(\vec{s})} \pdv{\overline\Psi(\vec{s})}{\theta(\vec{r})} \right),
\end{align}
also evaluated at $\beta=0$. These relations describe how variations in $\beta$ or $\theta$ propagate through steady-state fields via the system’s linearized response near equilibrium.

Using Eq.~\eqref{eq:app_stat_cond}, we expand Eq.~\eqref{eq:app_delta_V2}, exchange integration order, and regroup terms by their $\beta$ derivatives
\begin{align}\label{eq:app_delta_V3}
\pdv{ C}{ \theta(\vec{r})} &= \ii\int \dd\vec{u} \int \dd\vec{s} \nonumber \\
&\Biggl[ \left( \pdv{ \overline\kappa^\beta(\vec{s})}{ \Psi^\beta(\vec{u})} \pdv{\Psi(\vec{s})}{\theta(\vec{r})} - \pdv{\kappa^\beta(\vec{s})}{ \Psi^\beta(\vec{u})} \pdv{\overline\Psi(\vec{s})}{\theta(\vec{r})} \right) \pdv{\Psi^\beta(\vec{u})}{\beta} \nonumber \\
&+\left( \pdv{\overline \kappa^\beta(\vec{s})}{ \overline\Psi^\beta(\vec{u})} \pdv{\Psi(\vec{s})}{\theta(\vec{r})} - \pdv{\kappa^\beta(\vec{s})}{\overline \Psi^\beta(\vec{u})} \pdv{\overline\Psi(\vec{s})}{\theta(\vec{r})} \right) \pdv{\overline\Psi^\beta(\vec{u})}{\beta}\Biggr].
\end{align}

Assuming that derivatives of $\overline{\kappa}^\beta$ with respect to $\Psi^\beta$ are invariant under spatial transposition and that the near-equilibrium conditions hold,
\begin{align} \label{eq:app_condition}
\pdv{\overline \kappa^\beta (\vec{s})}{\Psi^\beta (\vec{u})} = \pdv{\overline \kappa^\beta (\vec{u})}{\Psi^\beta (\vec{s})}, \qquad
\pdv{\kappa^\beta(\vec{s})}{ \Psi^\beta(\vec{u})} \approx - \pdv{ \overline \kappa^\beta(\vec{u})}{ \overline\Psi^\beta(\vec{s})},
\end{align}
comparison of Eqs.~\eqref{eq:app_stat_cond2} and \eqref{eq:app_delta_V3} gives
\begin{align}
\pdv{ C}{ \theta(\vec{r})} &\approx -\ii\int \dd\vec{u} \left( \pdv{ \overline\kappa^\beta(\vec{u})}{\theta(\vec{r})} \pdv{\Psi(\vec{u})}{\beta} - \pdv{\kappa^\beta(\vec{u})}{\theta(\vec{r})} \pdv{\overline\Psi(\vec{u})}{\beta}\right), \label{eq:app_delta_V_final}
\end{align}
evaluated at $\beta=0$.

\textit{Gross-Pitaevskii Equation.}
Returning to the polariton GPE, we now write the update rule explicitly. Condition~\eqref{eq:app_condition} holds when the imaginary part of the potential $V(\vec{r})$ and the dissipation parameter $\gamma$ are negligible (near-equilibrium regime), since the Liouvillian is spatially symmetric and the diagonal Jacobians are
\begin{align} \label{eq:app_gpe_jacob}
\pdv{\kappa^\beta(\vec{u})}{\Psi(\vec{s})} &= \left(\frac{ \ii \hbar}{2m}\nabla^2 -\ii\frac{V}{\hbar} -\ii\frac{2g}{\hbar}|\Psi(\vec{u})|^2 -\frac{\gamma}{\hbar}\right)\delta(\vec{s}-\vec{u}), \\
\pdv{\overline \kappa^\beta(\vec{u})}{\overline\Psi(\vec{s})} &= -\left(\frac{ \ii \hbar}{2m}\nabla^2 -\ii\frac{\overline V}{\hbar} -\ii\frac{2g}{\hbar}|\Psi(\vec{u})|^2 +\frac{\gamma}{\hbar}\right)\delta(\vec{s}-\vec{u}),
\end{align}
where $\delta(\vec{s}-\vec{u})$ is the Dirac delta function. For an arbitrary nonlinearity $f(\Psi)$, condition~\eqref{eq:app_condition} remains valid when
\begin{align}
f(\Psi)+ \Psi \pdv{f(\Psi)}{\Psi} = \overline{f}(\Psi) + \overline{\Psi} \pdv{\overline{f}(\Psi)}{\overline{\Psi}}.
\end{align}
%For real $V(\vec{r})$ and pump weight $w(\vec{r})$, outlined derivation yields Eqs.~\eqref{eq:delta_V_polariton}–\eqref{eq:delta_w_polariton}.

\textit{Discrete variable case.}
In the discretized system, the derivation simplifies. Define the state vector $s=(\Psi_i,\overline\Psi_i)$ and evolution vector $K=(-\ii \kappa^\beta_i, \ii \overline\kappa^\beta_i)$. Using Eq.~\eqref{eq:app_cost_diff2} and $K(s,\theta,\beta)\equiv0$, we find
\begin{align}
\Delta \theta_i &\propto - \pdv{C}{\theta_i} = - \pdv{C}{s_j}\pdv{s_j}{\theta_i} = \pdv{\overline K_j}{\beta}\pdv{s_j}{\theta_i}= -\pdv{\overline K_j}{s_k}\pdv{s_k}{\beta}\pdv{s_j}{\theta_i}.\nonumber
\end{align}
Under transposition invariance and the near-equilibrium condition, which now have the compact form
\begin{align} \label{eq:nep_dis}
\pdv{\overline K_j}{s_k} \approx \pdv{\overline K_k}{s_j},
\end{align}
and using again $K(s,\theta,\beta)\equiv0$, we obtain
\begin{align}
\Delta \theta_i &\propto
- \pdv{\overline K_k}{s_j}\pdv{s_k}{\beta}\pdv{s_j}{\theta_i}
= \pdv{\overline K_k}{\theta_i}\pdv{s_k}{\beta} \nonumber \\
&= i\pdv{\overline{\kappa}_k}{\theta_i}\pdv{\Psi_k}{\beta}
- i\pdv{\kappa_k}{\theta_i}\pdv{\overline{\Psi}_k}{\beta}.
\end{align}

\section{Comparison with the vector-field Equilibrium Propagation} \label{appB}

In this Appendix, we clarify the difference between the NEP training principle and the vector-field equilibrium-propagation result of Scellier et al.~\cite{Scellier2018VectorField}, applied separately to the real and imaginary parts of the wave function.

A direct application of the update rule of Ref.~\cite{Scellier2018VectorField} to the real and imaginary parts of the wave function does not lead to a useful update rule for complex wave systems. To demonstrate this, we provide below a step-by-step application of the vector-field equilibrium-propagation scheme in the real and imaginary basis.

We consider the dimensionless GPE as a simple example of a nonlinear wave equation,
\begin{align}
    i\partial_t \Psi = \left(-\nabla^2+V+f(\Psi)-i\gamma\right)\Psi+iP.
\label{eq:free_gpe_appendix}
\end{align}
In the real and imaginary basis, it reads
\begin{align}
    \partial_t \begin{pmatrix} \psi_R\\ \psi_I \end{pmatrix} &= \begin{pmatrix} \kappa_R\\ \kappa_I \end{pmatrix} \nonumber\\
    &= \begin{pmatrix}
    -\gamma\psi_R +\left(-\nabla^2+V+f(\psi)\right)\psi_I +P_R \\
    -\gamma\psi_I -\left(-\nabla^2+V+f(\psi)\right)\psi_R +P_I
    \end{pmatrix}, \label{eq:gpe_real_imag_appendix}
\end{align}
where we assumed $V\in\mathbb{R}$. Notice the sign difference between the real and imaginary mixing terms. This structure is characteristic of wave equations and differs from the symmetric coupling matrix assumed in Ref.~\cite{Scellier2018VectorField}.

We denote $s=(\psi_R,\psi_I)$, $K=(\kappa_R,\kappa_I)$. The condition in Ref.~\cite{Scellier2018VectorField},
\begin{align}
    \frac{\partial C}{\partial s} = -\frac{\partial K}{\partial\beta}
\end{align}
leads to the nudging prescription
\begin{align}
    \kappa_R^\beta = \kappa_{0R} -\beta\frac{\partial C}{\partial\psi_R}, \qquad
    \kappa_I^\beta = \kappa_{0I} -\beta\frac{\partial C}{\partial\psi_I}.
    \label{eq:nudge_scellier_appendix}
\end{align}

Following the corresponding derivation in Ref.~\cite{Scellier2018VectorField}, the update rule requires a symmetric Jacobian matrix, 
\begin{equation}
    \frac{\partial K}{\partial s} = \left(\frac{\partial K}{\partial s}\right)^T,
\end{equation}
or, explicitly,
\begin{equation}
    \begin{pmatrix}
    \dfrac{\partial\kappa_R}{\partial\psi_R} & \dfrac{\partial\kappa_R}{\partial\psi_I} \\
    \dfrac{\partial\kappa_I}{\partial\psi_R} & \dfrac{\partial\kappa_I}{\partial\psi_I}
    \end{pmatrix} = \begin{pmatrix}
    \left(\dfrac{\partial\kappa_R}{\partial\psi_R}\right)^T & \left(\dfrac{\partial\kappa_I}{\partial\psi_R}\right)^T \\
    \left(\dfrac{\partial\kappa_R}{\partial\psi_I}\right)^T &
    \left(\dfrac{\partial\kappa_I}{\partial\psi_I}\right)^T
    \end{pmatrix}.
\end{equation}
However, due to the sign difference between the real and imaginary parts of the
equation,
\begin{align}
    \frac{\partial\kappa_R}{\partial\psi_I} \neq \left( \frac{\partial\kappa_I}{\partial\psi_R} \right)^T,
\end{align}
and therefore
\begin{align}
    \frac{\partial K}{\partial s} \neq \left( \frac{\partial K}{\partial s} \right)^T,
\end{align}
even when the evolution equation contains only symmetric differential operators, such as the Laplacian.

By contrast, in the NEP training scheme, the nudging can be written in the real and imaginary basis as
\begin{align}
    \kappa_R^\beta = \kappa_{0R} +\beta\frac{\partial C}{\partial\psi_I}, \qquad
    \kappa_I^\beta = \kappa_{0I} -\beta\frac{\partial C}{\partial\psi_R}. \label{eq:nudge_nep_appendix}
\end{align}
This prescription is different from Eq.~\eqref{eq:nudge_scellier_appendix} and results in a symmetric Jacobian matrix in the case of a conservative wave equation.

Physically, in contrast to the original vector-field equilibrium-propagation prescription of Ref.~\cite{Scellier2018VectorField}, the output is not nudged directly toward the target. Instead, the real part is nudged according to the difference between the imaginary part of the output and the corresponding target, whereas the imaginary part is nudged in the opposite direction according to the difference between the real part of the output and its target. In other words, each component is nudged according to the error associated with its conjugate variable. 

The NEP nudging prescription therefore physically and mathematically differs from the vector-field equilibrium-propagation rule introduced in Ref.~\cite{Scellier2018VectorField}.

\section{Additional Simulation Results.} \label{appC}
\begin{figure}[hbtp]
\centering
\includegraphics[width=0.48\textwidth]{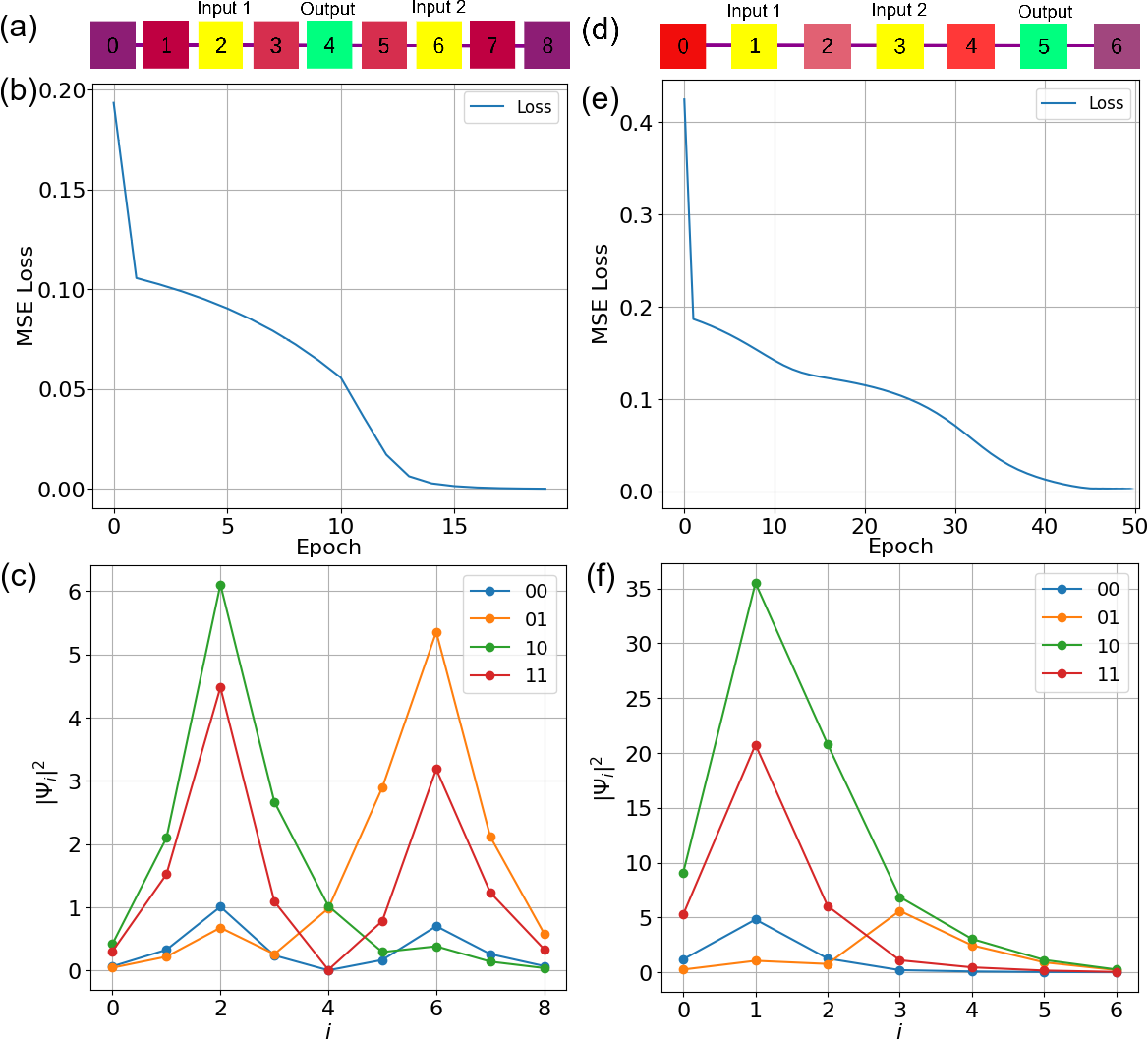}
\caption{\justifying XOR task in (a) a 9-node 1D network with training restricted to $V$, fixed near-optimal pump weights $w$, and (d) a 7-node network with asymmetric input placement (sites 1, 3) and output at site 5. Top panels: training losses (b) for (a) and (e) for (d). Bottom: steady-state field magnitudes $|\Psi|^2$ for each input (c) and (f). (a)–(c) show that $V$--only training solves XOR for near-optimal fixed $w$; (d)–(f) show successful training despite broken spatial symmetry.}
\label{fig:app_xor_app}
\end{figure}

\textit{XOR task.} Figure~\ref{fig:app_xor_app}(a)–(c) shows XOR training on the same 9-node architecture as in the main text, but with optimization limited to the potential $V$. Pump weights $w$ are fixed near optimal ($w_2=1$, $w_6=-0.75$), serving as static input encoding. Even under this restriction, the network minimizes loss and reproduces correct outputs, confirming that proper input encoding via $w$ enables $V$-only training, \change{though convergence saturates later than with both $V$ and $w$ optimized}. Figure~\ref{fig:app_xor_app}(d)–(f) presents a reduced 7-node network with asymmetric input placement (sites 1 and 3) and output at site 5, where both $V$ and $w$ are trained. Results show that spatial symmetry of inputs is unnecessary—the network still performs the XOR mapping under asymmetric geometry. This illustrates NEP’s robustness to architectural variation.

\begin{figure}[hbtp]
\centering
\includegraphics[width=0.48\textwidth]{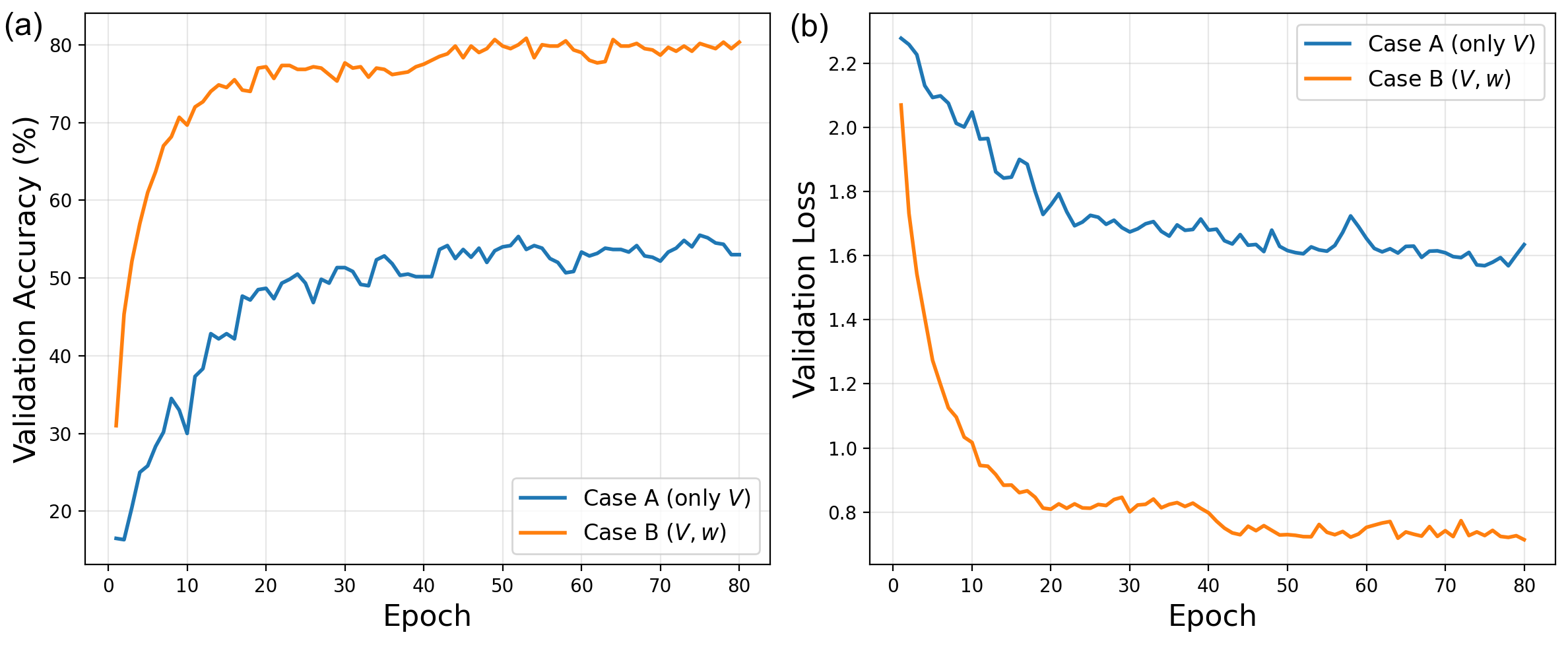}
\caption{\justifying Impact of the choice of trainable parameters, $V$ and $w$, on (a) validation accuracies and (b) CCE losses versus epochs during training of a $5\times50$-node 2D network on MNIST using NEP for 50 epochs. Classification achieves $\approx80\%$ accuracy when both $V$ and $w$ are trainable, and $\approx53\%$ when $w$ is fixed to unities.}
\label{fig:app_mnist_10}
\end{figure}

\textit{MNIST task.} Figure~\ref{fig:app_mnist_10} presents classification of the MNIST dataset (\change{after Principal Component Analysis}) using a $5\times50$ polariton grid with ten $5\times5$ input cells and ten output nodes. Training both $V$ and $w$ yields $\approx80\%$ test accuracy, while training only $V$ reduces accuracy to $\approx53\%$, still demonstrating purely optical optimization.

\end{document}